\documentclass[journal]{IEEEtran}

\usepackage{amsmath,amssymb}
\usepackage{graphicx}
\usepackage{cite}
\usepackage{array}
\usepackage{booktabs}
\usepackage{xcolor}
\usepackage{url}
\usepackage{tabularx}
\usepackage{hyperref}

\usepackage{algorithm}
\usepackage{algorithmic}

\usepackage[font=footnotesize,labelfont=bf]{caption}

\renewcommand{\arraystretch}{1.05}

\usepackage{enumitem}
\setlist[itemize]{leftmargin=*,nosep,topsep=1pt}

\begin{document}
\title{\huge Agentic AI as a Network Control-Plane Intelligence Layer for Federated Learning over 6G}
\author{Loc X. Nguyen, Ji Su Yoon, Huy Q. Le, Yu Qiao, Avi Deb Raha, Eui-Nam Huh,~\IEEEmembership{Member,~IEEE},\\ Nguyen H. Tran,~\IEEEmembership{Senior Member,~IEEE}, Zhu Han,~\IEEEmembership{Fellow,~IEEE}, Choong Seon Hong,~\IEEEmembership{Fellow,~IEEE} 

\IEEEcompsocitemizethanks{
\IEEEcompsocthanksitem Loc X. Nguyen, Huy Q. Le, Avi Deb Raha, Eui-Nam Huh, and Choong Seon Hong are with the Department of Computer Science and Engineering, Kyung Hee University, Yongin-si, Gyeonggi-do 17104, Republic of Korea. Emails: \{xuanloc088; quanghuy69; avi; johnhuh; cshong\}@khu.ac.kr.
\IEEEcompsocthanksitem Ji Su Yoon, and Yu Qiao are with the Department of Artificial Intelligence, Kyung Hee University, Yongin-si, Gyeonggi-do 17104, Republic of Korea. Email: \{yjs9512, qiaoyu;\}@khu.ac.kr.
\IEEEcompsocthanksitem Nguyen H. Tran  is with the School of Computer Science, The University of Sydney, Australia, (email: nguyen.tran@sydney.edu.au)
\IEEEcompsocthanksitem Zhu Han is with the Electrical and Computer Engineering Department, University of Houston, Houston, TX 77004, and also with the Department of Computer Science and Engineering, Kyung Hee University, Yongin-si, Gyeonggi-do 17104, Rep. of Korea, (e-mail: hanzhu22@gmail.com)

}
}



\maketitle

\begin{abstract}
The shift toward user-customized on-device learning places new demands on wireless systems: models must be trained on diverse, distributed data while meeting strict latency, bandwidth, and reliability constraints. To address this, we propose an Agentic AI as the control layer for managing federated learning (FL) over 6G networks, which translates high-level task goals into actions that are aware of network conditions. Rather than simply viewing FL as a learning challenge, our system sees it as a combined task of learning and network management. A set of specialized agents focused on retrieval, planning, coding, and evaluation utilizes monitoring tools and optimization methods to handle client selection, incentive structuring, scheduling, resource allocation, adaptive local training, and code generation. The use of closed-loop evaluation and memory allows the system to consistently refine its decisions, taking into account varying signal-to-noise ratios, bandwidth conditions, and device capabilities. Finally, our case study has demonstrated the effectiveness of the Agentic AI system's use of tools for achieving high performance.

\end{abstract}


\section{Introduction}
Large language models (LLMs) have demonstrated their high capabilities across a range of complex scenarios, including user chatbots, code generation, and math solving~\cite{qiao2025deepseek}. These specialized models can be referred as LLM-based agents, and they require extensive human engineering to operate within a predefined environment and objective, and typically follow the rule-based designs. This property limits deployments in dynamic real-world environments, which require high adaptation to changes. Therefore, the Agentic AI has been developed to provide persistent, high-level autonomy and can orchestrate extended sequences of actions, typically in open-ended, real-world environments.

Agentic AI systems have emerged as a promising paradigm for autonomous control and optimization in 6G networks, spanning open radio access networks \cite{rezazadeh2025rivaling}, multi-objective network optimization \cite{chergui2025tutorial}, and high-flexibility network management \cite{11052733}. In this work, we extend this vision by positioning Agentic AI as a control-plane intelligence layer for federated learning (FL), orchestrating distributed client participation and managing data-plane transmissions across the wireless network. It is worth noting that FL can be employed as a foundation for training Agentic AI systems. Nevertheless, this paper focuses on the use of Agentic AI for FL design.


FL has been introduced \cite{mcmahan2017communication} as a distributed learning framework that leverages private data from local devices to train a deep learning (DL) model for a predefined task without centralizing the data. Specifically, in the FL framework, the server sends a model to clients, who train it on their local data and return the updates, and then the server aggregates these updates to improve the global model. The framework has been shown to be effective and deployed across multiple scenarios, including classification \cite{mcmahan2017communication}, sparse code multiple access \cite{10473705}, AI-generated content \cite{10398264}, and semantic communication \cite{11263916}.


However, these current frameworks are extremely costly because they require substantial engineering effort not only to meet the designed solution for a narrow, specific problem, but also to address the problem constraints. For instance, in a classification task, a wide range of problems must be addressed, including user data heterogeneity, domain shift, or limited communication/computation resources. Each of these challenges often necessitates the careful design and tuning of individual or combined mechanisms, such as client-selection policies, local training strategies, and global model aggregation rules, which substantially increases the complexity and engineering cost of FL system design.

\begin{table}[t]
\centering
\caption{Key difference between the two systems being operated by the traditional Agent and Agentic AI.}
\renewcommand{\arraystretch}{1.1}
\begin{tabular}{|p{1.3cm}|p{2.3cm}|p{3.8cm}|}
\hline
\textbf{Dimension} & \textbf{Traditional Agents} & \textbf{Agentic AI} \\ \hline

Task Scope &
Narrow, domain-specific tasks &
Broad, cross-domain capabilities (coding, analysis) \\ \hline

Core Capability &
Execute predefined rules &
Generate plans, reason, and adjust module/plans autonomously \\ \hline

Flexibility &
Low: fixed by hand-crafted logics &
High: learn new tasks with multi-collaboration of agents  \\ \hline

Autonomy Level & Requires explicit instructions &
Proactive: Planning multi-step actions, self-reflective learning \\ \hline

Learning Ability &
No inherent learning & Learns from context, memory, interaction, and agent feedback \\ \hline

Reasoning &
Rule-based or symbolic reasoning &
Natural-language reasoning, tool use, and chain-of-thought \\ \hline

Unseen Tasks & Fail if requiring outside predefined rules and logic &
 Actively searching for additional information and instruction, and able to learn on the fly  \\ \hline

Safety Risks &
Low due to the predictable action, bounded by rules &
Extreme high due to hallucinations, misalignment, and dual-use concerns of Agentic AI system \\ \hline

\end{tabular}
\label{Table1}
\end{table}

Motivated to address this bottleneck, we position Agentic AI as an intelligent control plane that jointly optimizes learning performance, wireless communication efficiency, and network resource allocation in dynamic wireless environments. Specifically, we first decompose the FL process into multiple steps, in which the Agentic provides decisions, such as client selection, wireless resource allocation and scheduling, and communication compression to improve spectrum efficiency and secure an optimization framework tailored to the given task. Unlike the study \cite{li2025position}, which provides only a brief discussion of the synergy between the two techniques without specifying a detailed approach to improve the FL framework. The contributions of the studies can be summarized as follows:
\begin{itemize}
    \item Our work starts by reframing the FL workflow as an autonomous decision-making system. By disentangling it into a series of interdependent stages, we create a foundation upon which Agentic AI can systematically design, refine, and manage each step.
    \item Secondly, we outline the fundamental building blocks of an Agentic AI system and highlight how these components can improve the robustness and enable the customization to design a task-oriented FL framework. Specifically, we emphasize the importance of task decomposition and specialized agent collaboration, which enable system evolution.
    \item Finally, we delve into the reasoning and the planning abilities of agents and show how they strengthen the Agentic AI system. Their integration serves as a reinforcing mechanism that enables FL to autonomously architect, optimize, and train learning models with minimal human oversight.
\end{itemize}

\section{Agentic AI and Federated Learning}
\subsection{Agentic AI Principles and Components}
\subsubsection{Principles}
Current classical Agent-based systems are specifically designed to perform a set of pre-defined tasks by following pre-defined rules, which restrict the system's autonomy, narrow the scope of tasks, and require human intervention for rule setup \cite{SAPKOTA2026103599}. To overcome these limitations of AI agents, an emerging system called Agentic AI has been introduced to extend their capabilities, as shown in Table~\ref{Table1}. Evolving from the traditional one, the Agentic AI has the abilities of self-learning, generative reasoning, and adaptation to the environment/unstructured input, which enable a more autonomous system and make it easier to meet the demands of personal-oriented tasks \cite{11152698}.
\subsubsection{Components}
These advancements are achieved through four main components: \textit{1) Multi-agent collaboration}, \textit{2) Goal Decomposition}, \textit{3) Multi-step reasoning and Planning}, \textit{4) Reflective Reasoning and Memory Capabilities}. Each component individually contributes and complements the others to build a layered, self-governing system, where high-level objectives are iteratively translated into coordinated, measurable action.
\begin{figure}[t]
\centering
\includegraphics[width=0.5\textwidth]{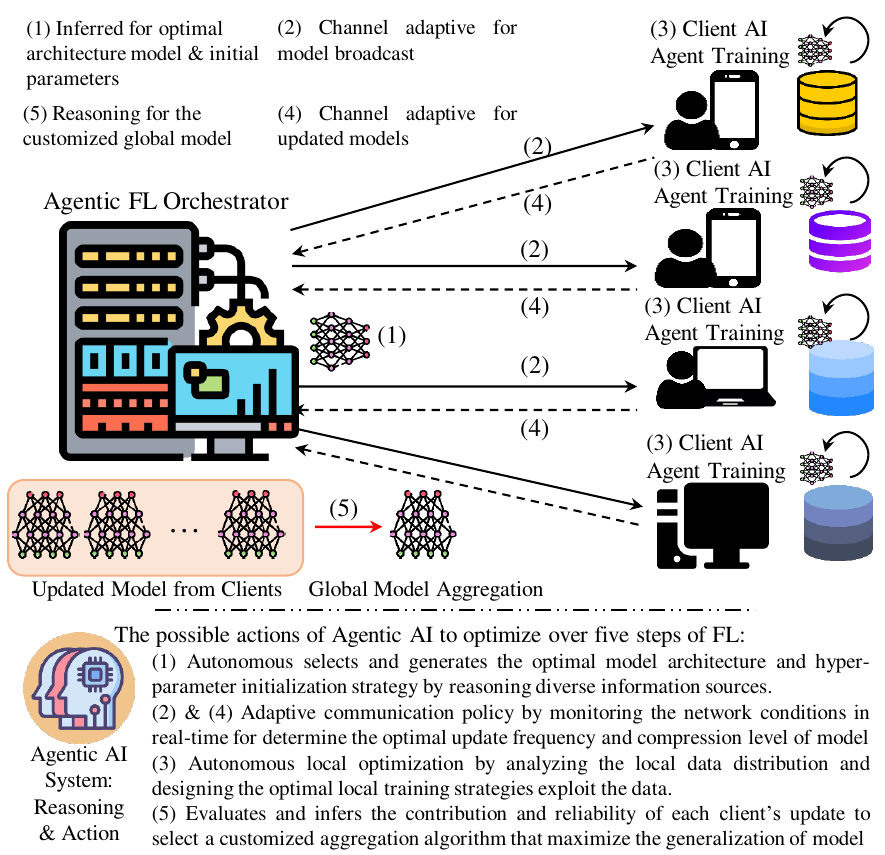}
\caption{We provide an overview of the integration of the Agentic AI system into the FL framework, where we outline possible actions at each step to secure strong generalization performance and adapt to channel conditions.}
\label{FLOverall}
\end{figure}
\begin{itemize}
    \setlength{\leftskip}{-0.01cm}
    \item \textbf{Multi-agent Collaboration:} Evolving beyond single-agent or traditional multi-agent setups, Agentic AI emphasizes the coordinated collaboration of multiple specialized agents. By assigning agents distinct roles and expertise, the system actively avoids functional overlap, minimizes unnecessary computation, and prevents inter-agent interference, thereby conserving inference resources and streamlining execution. For example, under the Agentic AI system, we can have one agent for the information retriever for the current task, one agent for coding, and another agent for evaluating the system's performance and efficiency. 
    
    \item \textbf{Goal Decomposition:} The system breaks the overall task into a series of smaller steps, each of which can be assigned to a specialized agent for execution and refinement. By structuring the workflow this way, the system can handle complex tasks autonomously and no longer relies on human intervention.
    
    \item \textbf{Multi-step Reasoning and Planning:} Breaking a complex goal into sub-tasks naturally expands the choices the system has to make, adds more moving parts, and increases the chance that it might drift away from what it is supposed to achieve. To manage the increasing decision space, Agentic AI uses multi-step reasoning and planning. By checking the intermediate reward at each step and considering how the current action affects future actions, the system can envision the future path through its reasoning and planning, and avoid veering off course.
    
    \item \textbf{Reflective Learning and Memory Capabilities:} A key advancement in Agentic AI is its ability to retain, revisit, and learn from the system's history. It structurally stores memory related to multi-step decisions, inter-agent exchanges, and, finally, the associated performance across different scenarios. These experiences form an active knowledge base that the Agentic AI can analyze, reinterpret, and reflect on learning when it encounters a new task. Through this reflective process, the system can recognize patterns in its behavior, assess the impact of prior actions, and determine which strategies were effective.
\end{itemize}

\subsection{Federated Learning}
\subsubsection{Basic Workflow} The original FL framework starts with the model that needs to be trained with a large amount of data. First, the learning model's parameters are initialized on the server and then distributed to the clients for training. Here, clients leverage their private data to update the parameters of the received model, thereby significantly improving the learning model's performance. Afterward, the server aggregates a global model from the models received from clients to combine the knowledge learned independently by each client and achieve generalization performance. The aggregated model is then broadcast to the client for the next training round, and this process continues until the learning model secures high performance. We illustrate the complete FL process shown in Fig.~\ref{FLOverall}.

\subsubsection{Interpretable Processes} Various studies have demonstrated average performance under the standard FL framework, which has motivated them to propose variants to improve the model's learning ability or low  in the distributed learning paradigm. Most of the studies focus on three techniques: \textit{1) Local training enhancements}, \textit{2) Global Aggregation Strategies}, and finally \textit{3) Computation and Communication Efficiency}. 

\begin{itemize}
    \setlength{\leftskip}{-0.01cm}
    \item \textbf{Local Training Enhancements:} The idea is relatively simple and direct, with the objective of enhancing the global model's learning capability; we first strengthen the training process of each individual client within the network. Consequently, the aggregated global model from a set of improved models eventually achieves higher overall effectiveness. 
 
    \item \textbf{Global Aggregation Strategies:} The global model aggregation has a significant impact on its performance, which can be both negative or positive. For example, we want to eliminate fake clients created by an attacker in the network from contribute in the aggregation process. On the other hand, the server can assign greater weight to clients with high-quality data than to those with below-average-quality data to achieve more robust and consistent results \cite{FLAggregation}.
    \item \textbf{Computation and Communication Efficiency:} Approaching the FL framework from a different narrative, where the clients are constrained by various factors, including computation and communication resources \cite{10856890}. To address this, the research community has developed a range of techniques, including model sparsification and quantization, to reduce communication complexity.
\end{itemize}

\begin{figure*}[t]
\centering
\includegraphics[width=1\textwidth]{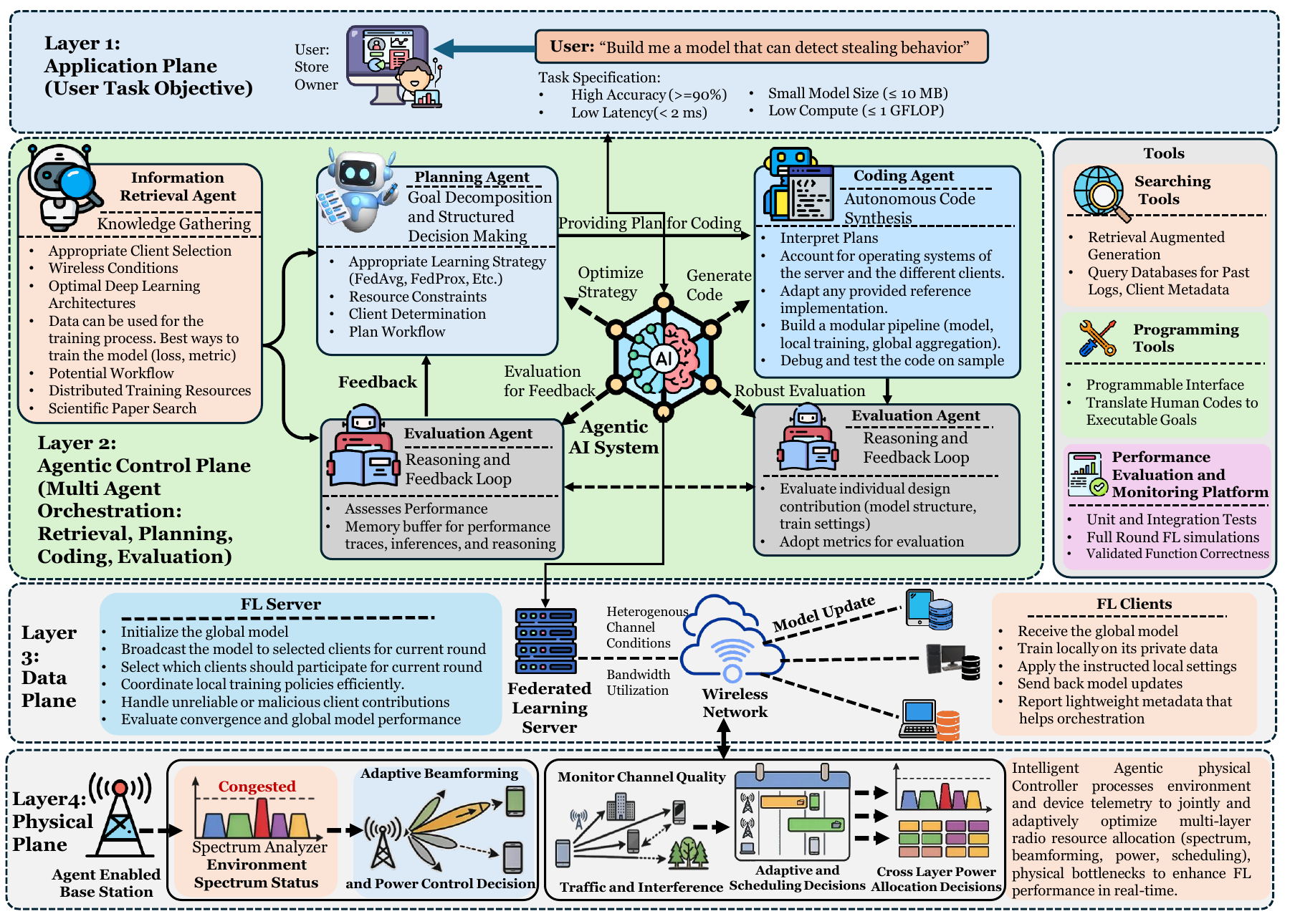}
\caption{The illustration of the autonomy of the Agentic AI system to train a deep learning model for a user-oriented task with the FL frameworks. Our proposed system is human-free, with specialized agents actively collaborating to train for the task.}
\label{Proposal}
\end{figure*}

\section{Integration of Agentic AI for Federated Learning Workflow}
The adoption of the Agentic AI system to autonomously design an FL framework for a specific task without human intervention is a natural and practically inevitable evolution toward scalable, adaptive, and user-oriented learning systems. In this section, we discuss how each special property of the Agentic AI architecture can significantly enhance different stages of the FL process and the overall system is shown in Fig~\ref{Proposal}.

\subsection{Planning Property: FL from the Viewpoint of Agentic AI}

The planning capability of the Agentic AI system, supported by goal decomposition, is essential for achieving autonomous, structured decision-making throughout the FL process. While most FL research focuses on model optimization, practical deployments must simultaneously manage network-centric concerns such as client selection, incentive mechanisms, scheduling, and wireless resource allocation (bandwidth, power, and latency). The Agentic AI system must therefore interpret the decisions at multiple, interdependent stages: \textit{1) task objective specification}, \textit{2) client selection \& incentive design}, \textit{3) local training and server aggregation}, and \textit{4) wireless communication and resource management}. In this role, planning connects high-level learning objectives with practical, network-aware decisions, allowing the system to turn abstract goals into concrete actions that account for both model performance and network constraints.
\subsubsection{Task Objective}
A well-defined task objective guides not only learning targets but also the network trade-offs the system is willing to accept (e.g., accuracy vs. latency or communication cost). The retrieval agent in the Agentic AI system first converts a user’s task specification into a flexible objective that can evolve during training. The planning agent then optimizes that objective by refining its strategy while considering the wide range of factors: user’s accuracy and latency requirements, feedback from the evaluation agent, client compute and energy budgets, and current network conditions (e.g., available uplink bandwidth, channel quality).

\subsubsection{Client Selection and Incentive Design}

Eliminating human intervention in the FL process requires the Agentic AI system to autonomously select appropriate clients from the network to participate in training. This stage is critical to task performance, as the data held by selected clients must align with the current learning objective. To resolve this, the Agentic AI system incorporates the client data and resource information when prioritizing participants, favoring clients that provide both relevant data and stable computing capacity. In addition, an incentive mechanism can be designed by the intelligent system to encourage the client to participate in the learning process and compensate them for the computing resources.

\subsubsection{Local Training and Server Aggregation}

The agent can establish the client's status by collecting its telemetry and historical log data statistics; after which, the planning agent generates a training plan tailored to each client's data characteristics and resource constraints. Critically, it can actively adjust hyperparameters such as the learning rate, batch size, and number of epochs using optimization tools that recommend a suitable training strategy. 
Through this layered and context-aware reasoning, the planning agent transforms local training into a structured, adaptive process that aligns with the global learning objective. Upon receiving client updates, the system performs global aggregation using historical patterns stored in memory. The evaluation agent then assesses model efficiency and determines how updates should be weighted, filtered, or combined, including applying domain-aware rules and discarding unreliable contributions. Finally, convergence statistics and global performance metrics are recorded to refine future decisions. Through this continuous loop of retrieval, reasoning, and memory updates, the server guides the FL process in a stable and adaptive manner.

\subsubsection{Wireless Communication and Resource Management}

In distributed FL over wireless networks, clients experience heterogeneous channel conditions, traffic loads, and device capabilities, leading to uplink congestion, variable latency, and straggler effects that reduce the system throughput \cite{10239348}. To address this, an agent-enabled base station operates as an intelligent control-plane entity that continuously models uplink congestion based on a wide range of factors: queue states, interference levels, and bandwidth utilization. Therefore, the agent jointly optimizes the resource allocation, transmission power, and scheduling priority to balance special efficiency and fairness.

\subsection{Collaboration Property among Multi-Agent of Agentic AI}

As stated above, the Agentic AI system is responsible for making decisions across several steps, including the DL architecture for the task, potential clients, incentive mechanism design, local training algorithm, aggregation technique, resource management for communicating the model within the network, and even generating the code for the entire learning process. This is a challenging problem for a conventional AI agent, which only contains a single agent. On the other hand, Agentic AI consists of multiple specialized agents: an information-retrieval agent, a planning agent, an evaluation/reasoning agent, and, finally, a coding agent that can collaborate with the others to sufficiently complete highly complex tasks. Their role in the FL process is as follows:

\textbf{The information-retrieval agent} is responsible for gathering information relevant to the FL task, such as clients with appropriate data in the network, wireless conditions for communication, the intermediate status of the distributed training resources, the optimal DL architecture for the considered task, and the potential of optimal training workflow. The retrieved information is obtained by the agent from multiple sources, such as reports from the base station for wireless conditions, scientific papers for training strategy, and statistics on data distribution for the client information. The information is transferred to the planning agent to design the optimal system.

\textbf{The planning agent} takes the information from the information-retrieval agent into consideration and comes up with an appropriate learning strategy. This plan is required to address a wide range of requirements: the resource constraints of the wireless communication, client determination, training loss, and workflow. For example, it allocates more power to communicate with clients who possess high-quality data. In addition, the planning agent receives feedback from the evaluation agent, which explicitly indicates the inefficiency of system modules and adjusts the plan accordingly. 

\begin{table*}[t]
\centering
\caption{Comparison Between Conventional FL, AutoML for FL, and Agentic FL}
\label{tab:comparison}
\begin{tabular}{lccc}
\toprule
\textbf{Feature} & \textbf{Conventional FL} & \textbf{AutoML for FL} & \textbf{Agentic FL (Proposed)} \\
\midrule
Hyperparameter tuning & Manual & Automated & Multi-agent reasoning \\
Client selection & Heuristic & Static optimization & Context-aware planning \\
Network awareness & Limited & Minimal & Joint cross-layer reasoning \\
Control-plane intelligence & None & Limited & Persistent memory-based \\
\bottomrule
\end{tabular}
\end{table*} 

\textbf{The coding agent} receives plans from the planning agent, and it synthesizes the code from scratch to implement the whole FL pipeline, including the construct learning model architecture for the task, resource management for the wireless network, the training strategies at clients, and the aggregation method at the server. On the other hand, it can also adapt and refine existing implementations when reference code is provided, ensuring the system successfully reflects the intended design. In a real-world scenario, the client and server may use different languages; the code agent must actively convert the source code into executable code for each device.

\textbf{The evaluation agent} systematically assesses the contribution of each design decision to the system’s final performance after completion of coding and training framework. Specifically, the agent actively chooses the metric that satisfies the task requirements, and at multiple levels, deploys to evaluate the impact of every single step of the system on the final performance, and also stores the decision-making associated with the performance in a memory buffer. With its memory capabilities, the agent can reason about the performance of the learning process, such as which modules/steps are responsible for performance degradation or which steps contribute to improvement. This evaluation agent can provide detailed feedback to the planning agents to revise their learning strategies and steer the system toward user-oriented tasks.

\subsection{The Advantage of Tool Use Property in Agentic AI System}

The tool-use capability of the Agentic AI system significantly enhances its autonomy and practical effectiveness in FL deployments. External tools, such as data analytics modules, network monitoring software, and optimization solvers, enhance the agent's ability to ground its decisions in a real-time system. In particular, the Agentic AI system leverages optimization solvers to determine solutions to resource allocation problems in a dynamic environment and augments the server's aggregation mechanism using network reliability information from monitoring tools. Promising tools for the FL scenarios are:

\subsubsection{Searching Tools}

Searching tools enable the agent to retrieve, filter, and analyze essential information for informed decision-making during FL orchestration. By leveraging retrieval-augmented mechanisms, such as querying a retrieval-augmented generation database, the agent can access past training logs, client metadata, system states, and performance traces to support key decisions, including client selection, anomaly detection, resource estimation, and hyperparameter tuning. Additionally, the agent can conduct online searches across web resources to gather relevant information for decision-making in the FL process, such as local enhancement training techniques, global aggregation strategies to preserve the local knowledge, and model quantization to reduce communication resources.

\subsubsection{Programming Tools}

Another useful tool that the Agentic AI can benefit from is the programming tool, in which the Agentic AI can directly and autonomously translate human-level objectives into executable system operations. It actively generates, modifies, and validates the code in response to real-time feedback and environment change to rapidly adapt to the evolving network conditions, heterogeneous device constraints, and dynamic task requirements. In addition, the Agentic AI skips the human intervention by interacting directly with a programmable interface, shortening deployment cycles and ensuring operational consistency.

\subsubsection{Performance Evaluation and Monitoring Platform} provides an execution environment where the agent can autonomously run the generated code, verify functional correctness, and assess model performance without human intervention. After synthesizing the FL-related code modules, the agent deploys them into a controlled sandbox environment, where the evaluation engine systematically validates each component, including data-processing pipelines, local model updates, global aggregation procedures, and communication routines. Specifically, once the agent produces code for local training or global aggregation, it can immediately execute unit tests (e.g., checking whether gradients explode under certain client distributions), integration tests (e.g., verifying the consistency of global aggregation under simulated client dropouts), and full-round FL simulations (e.g., running FedAvg~\cite{mcmahan2017communication} on a CIFAR-10 partition to verify that accuracy improves monotonically).

\subsection{Benefits and Capabilities}
With the introduction of an Agentic AI system, FL becomes automated, enabling learning models to be autonomously designed and optimized in accordance with user-oriented requirements and scaled across a wide range of personalized tasks. By reducing dependence on human-crafted frameworks, the framework can lower engineering overhead, minimize human-induced bias, and alleviate downtime caused by manual intervention. Furthermore, the Agentic system can operate continuously, adapt to evolving tasks and environments, and refine learning strategies through iterative reasoning and feedback. As a result, it enhances both the scalability and robustness of FL, transforming it into a self-sustaining and user-adaptive learning paradigm. Table~\ref{tab:comparison} provides comparisons of our proposed Agentic AI against conventional FL and Automated Machine Learning (AutoML). Recent developments have introduced several practical Agentic AI systems, such as Anthropic’s Claude Cowork, a desktop assistant that helps users perform multi-step tasks using local tools; OpenClaw, an Agentic AI assistant capable of interpreting natural language instructions and autonomously making decisions; and MolBook Pro, a domain-specific agent designed for chemical and biochemical data management and analysis.

\section{Case Study: Agent Coding based on Retrieval Information}

\begin{figure*}[t]
\centering
\includegraphics[width=0.80\textwidth]{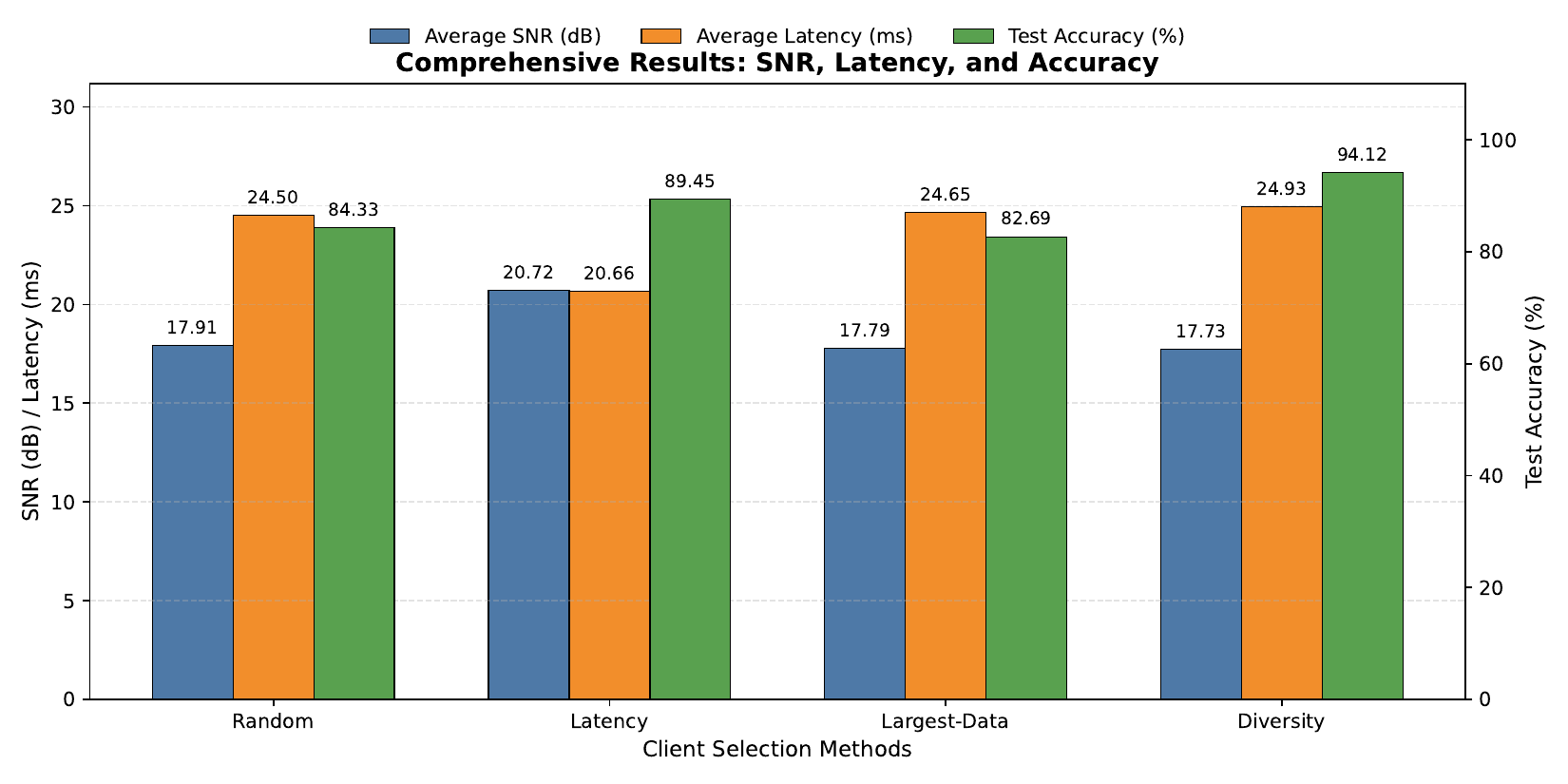}
\caption{Grouped bar chart comparing average SNR, average communication latency, and test accuracy across four client‑selection benchmarks (random, latency‑based, largest‑data, and class-diversity). The left y‑axis shows SNR/latency; the right y‑axis shows accuracy.}
\label{CaseStudyResult}
\end{figure*}
\subsection{Task Configuration}
Here, we consider a simple classification task on the MNIST dataset based on the online framework\footnote{https://github.com/shaoxiongji/federated-learning}. We explicitly model a bandwidth-constrained wireless network in which the central server is utilized for downlink transmission. The simulation configurations are summarized in Table~\ref{sim_tab}. Within the Agentic AI workflow:
\begin{itemize}
    \setlength{\leftskip}{0.1cm}
    \item GitHub Copilot serves as the Coding Agent.
    \item Visual Studio Code provides the execution and evaluation environment.  
    \item Leveraging diverse external tools (e.g., wireless channel measurement and client-level data analysis) to enhance autonomous decision-making performance.
\end{itemize}

\begin{table}[t]
\centering
\renewcommand{\arraystretch}{1.1}  
\caption{Simulation Parameters}
\label{sim_tab}
\begin{tabular}{|l|c|}
\hline
    \textbf{Constrained Resources} & \textbf{Value} \\ \hline \hline
     Number of Potential Clients & 15 \\ \hline
     Dropout Clients & 30\% \\ \hline
     Communication Bandwidth Per Channel  &  5 MHz  \\ \hline
     Number of Available Channels & $\{3,4,5,6,7\}$ \\ \hline
     SNR	Range for User &   10-25 dB \\ \hline
     Communication Rounds &   10 \\ \hline
     Dirichlet Distribution $\alpha$ & 0.1  \\ \hline 
\end{tabular}
\end{table}
\subsection{Performance Evaluation}

As shown in Fig.~\ref{CaseStudyResult}, we compare the performance of four benchmarks that select clients based on different criteria. It is worth noting that all benchmarks are coded by the Coding Agent with minimal human intervention, and we only provide directions on selecting clients. The latency-based selection leverages the SNR information from measurement tools, the largest-based selection utilizes the number of samples information from clients. On the other hand, class-diversity selection collects and analyzes data on client class diversity to identify the set of participants. The diversity-based approach achieves the highest testing accuracy, while the latency-based selection secures the lowest delay and the highest SNR. The results emphasize the high potential of Coding Agent in terms of tool usage and perceive the human instruction.

\section{Challenges \& Open Issues for Agentic AI in FL}
\subsection{Technical Challenges}
\textit{The stability of Learning-Control Coupling:} it requires us to pay close attention to issues like closed-loop instability, the impact of delayed feedback, and the challenges of tuning hyperparameters. When the Agentic AI functions as a control layer, it continuously receives learning metrics-such as loss, gradient variance, and convergence rate-to make adjustments to key control variables like client selection, learning rate, compression ratio, and update frequency. This creates a discrete-time feedback system where aggressive control settings or complex interactions between learning dynamics and network scheduling can lead to instability~\cite{11192484}. Additionally, delays in feedback-caused by factors like communication latency, slow participants, or asynchronous data aggregation-often result in decisions based on outdated information


\textit{Multi-Agent Game-Theoretic Conflict:} occurs when several specialized agents in the control-plane intelligence layer pursue goals that are not completely aligned while navigating shared network and learning constraints. For example, a planning agent might focus on achieving quick results by choosing clients with high data rates or signal quality, while a resource-management agent is trying to cut down on bandwidth use or energy consumption. At the same time, an incentive-design agent may work to lower the costs of participation, which can clash with the robustness requirements set by the evaluation agent. Therefore, it creates conflict among agents and affects the overall performance of the Agentic system. 

\textit{Scalability} becomes a significant challenge when implementing Agentic AI as a control-layer in large-scale FL systems that involve a vast number of clients and extensive operational histories. If we simply store telemetry and training logs for each client in every round, we end up with a memory complexity that scales with the product of the number of clients and the number of training rounds. Centralized decision-making adds to this by requiring inputs that also scale with the number of clients, which means that planning delays increase as we add more clients. Moreover, the back-and-forth communication between multiple agents can increase the messaging load, leading to potential congestion in the control layer and slower decision-making.

\subsection{Open Issues}

\textit{Misuse and Task-Safety Concerns:}
The introduction of the Agentic AI system offers an autonomous, human-free intervention, a user-oriented task for scalability, but also raises unprecedented safety considerations. As the system can automatically design models, generate code, and optimize training strategies, it becomes possible for users to unintentionally or deliberately direct the framework toward high-risk tasks, such as sensitive-attribute inference, unauthorized surveillance, or deepfake generation. Unlike traditional FL, where the workflow requires expert supervision, an agentic FL system lowers the barrier for constructing powerful models, increasing the risk of dual-use behavior. Ensuring that the agent operates within well-defined, safe task boundaries thus becomes a core challenge for future deployments.

\textit{Ethical Alignment and Value Drift Over Time:}
The Agentic AI is flexible to adapt to new user-oriented tasks and environments, which can encourage the internal decision patterns to drift away from safe and intended behavior. A proper, regular alignment check is required to prevent the agent from conflicting with organizational or regulatory expectations.

\textit{From Agentic AI to Physical AI Integration:}
Intelligent systems are increasingly transitioning from digital environments into the physical world, giving rise to physical AI-embodied systems capable of sensing, acting, and learning. This trend motivates the extension of Agentic AI from purely software-driven autonomy to embedded, embodied autonomy, where agents can actively coordinate physical actions on robots, drones, and smart devices. However, integrating agentic capabilities into physical platforms introduces an additional layer of complexity, as these systems must contend with real-world uncertainty, limited on-device resources, and safety-critical operational constraints.

\section{Conclusion and Future Research}
Our paper envisions the new paradigm in which an Agentic AI system autonomously designs, implements, and optimizes the FL framework pipeline for user-oriented tasks. By approaching the FL as a sequence of interpretable decision stages and allocating those stages to specialized agents for design (information retrieval, code synthesis, evaluation, and planning), we emphasize the future of human-engineering-free FL systems tailored to user-oriented tasks. This Agentic workflow accelerates deployment and supports scalable adaptation by leveraging the high-level goal decomposition and the coordinated collaboration among agents. In addition, we illustrate how multi-step reasoning and planning can be deployed to enhance the system's optimization ability, while the evaluation agent provides feedback to coordinated agents to steer the learning process, which plays a key role in the system. Finally, we identify key technical challenges, including the verification of auto-generated code and hallucination in LLM-based agents, and discuss open issues related to system misuse and long-term ethical alignment.

\bibliographystyle{IEEEtran}
\bibliography{IEEEabrv,ref_ieee_macros}


\section*{Biographies}
\footnotesize
\noindent \textbf{Loc X. Nguyen} is a Research Professor in the Department of Computer Science and Engineering, Kyung Hee University, Yongin-si, Gyeonggi-do 17104, Republic of Korea. His research interests include federated learning, semantic communication, and intelligent systems.

\noindent \textbf{Ji Su Yoon} is a PhD candidate at the Department of Computer Science and Engineering at Kyung Hee University, Yongin-si, Gyeonggi-do 17104, Republic of Korea. His research interests include federated learning and reinforcement learning.

\noindent \textbf{Huy Q. Le} is a Research Professor in the Department of Computer Science and Engineering, Kyung Hee University, Yongin-si, Gyeonggi-do 17104, Republic of Korea. His research interests include federated learning, cross-domain, and large language models.

\noindent \textbf{Yu Qiao} is a Research Professor in the Department of Computer Science and Engineering, Kyung Hee University, Yongin-si, Gyeonggi-do 17104, Republic of Korea. His research interests include federated learning, segment models, and large language models.

\noindent \textbf{Avi Deb Raha} is a PhD candidate at the Department of Computer Science and Engineering at Kyung Hee University, Yongin-si, Gyeonggi-do 17104, Republic of Korea. His research interests include federated learning, semantic communication, and beamforming.

\noindent \textbf{Eui-Nam Huh} is a Professor in Computer Science and Engineering at Kyung Hee University, Yongin-si, Gyeonggi-do 17104, Republic of Korea. His research interests include distributed edge and cloud computing.

\noindent \textbf{Nguyen H. Tran} is a Professor in the School of Computer Science, The University of Sydney, NSW 2006, Australia. His research interests include federated learning, semantic communication, and large language models.

\noindent \textbf{Zhu Han} is currently a John and Rebecca Moores Professor with the Department of Electrical and Computer Engineering and the Department of Computer Science, University of Houston, Houston, TX, USA. His research interests include wireless resource allocation and management, wireless communications and networking, quantum computing, data science, and privacy.

\noindent \textbf{Choong Seon Hong} is a Professor in the Department of Computer Science and Engineering, Kyung Hee University, Yongin-si, Gyeonggi-do 17104, Republic of Korea. His research interests include federated learning, wireless communication, and intelligent networking.

\end{document}